\documentclass[journal]{IEEEtran}

\ifCLASSINFOpdf 
\else
   \usepackage[dvips]{graphicx}
\fi
\usepackage{url}

\usepackage{graphicx}

\usepackage[dvipsnames]{xcolor}

\usepackage{cite}
\usepackage{algpseudocode}
\usepackage{algorithm}
\usepackage{amsmath}
\usepackage{multirow}
\usepackage{graphicx}
\usepackage[dvipsnames]{xcolor}
\usepackage{epstopdf}
\usepackage{caption}
\usepackage{subcaption}
\usepackage{bm}
\usepackage{pgfplots}
\usepackage{footnote}
\usepackage{stfloats}
\usepackage{placeins}
\usepackage{color,soul}
\usepackage{dsfont}
\usepackage{amssymb}
\usepackage{pbox}
\usepackage{enumerate}
\usepackage{etextools}
\usepackage{mathtools}
\usepackage{amsthm}
\usepackage{tabulary}
\usepackage{dsfont}
\usepackage{balance}

\usepackage[T1]{fontenc}

\usepackage[colorlinks = true]{hyperref}

\newtheorem{Remark}{Remark}[section]

\usepackage{tikz}
\usetikzlibrary{calc} 

\usepackage{cleveref}
\usepackage{comment}

\usepackage{wrapfig}

\def\I{{\mathbf I}}

\def\RR{{\mathbb R}}

\newcommand{\vct}[1]{\mathbf{#1}}
\newcommand{\mtx}[1]{\mathbf{#1}}

\newcommand{\calA}{\mathcal{A}}

\newcommand{\calH}{\mathcal{H}}

\newcommand{\calN}{\mathcal{N}}

\newcommand{\calX}{\mathcal{X}}
\newcommand{\calY}{\mathcal{Y}}

\newcommand{\vr}{\vct{r}}
\newcommand{\vs}{\vct{s}}

\newcommand{\vu}{\vct{u}}

\newcommand{\vx}{\vct{x}}
\newcommand{\vy}{\vct{y}}

\newcommand{\mA}{\mtx{A}}

\newcommand{\mS}{\mtx{S}}

\newcommand{\mU}{\mtx{U}}

\newcommand{\mX}{\mtx{X}}

\newcommand{\mOmega}{\mtx{\Omega}}

\def \st {\operatorname*{s.t.\ }}

\newcommand{\tensor}[1]{\boldsymbol{\mathcal{#1}}}
\def \tA {\tensor{A}}

\def \tS {\tensor{S}}

\def \tX {\tensor{X}}
\def \tY {\tensor{Y}}
\def \tZ {\tensor{Z}}

\begin{document}

\title{StoTAM: Stochastic Alternating Minimization for Tucker-Structured Tensor Sensing}

\author{Shuang Li, \IEEEmembership{Member, IEEE}
\thanks{Manuscript created January, 2026. 
This work was supported in part by NSF grants ECCS-2409702. This work has been submitted to the IEEE for possible publication. Copyright my be transferred without notice, after which this version may no longer be accessible.}
\thanks{Shuang Li (lishuang@iastate.edu) is with the Department of Electrical and Computer Engineering, Iowa State University, Ames, Iowa 50014 USA.}
}

\maketitle

\begin{abstract}

Low-rank tensor sensing is a fundamental problem with broad applications in signal processing and machine learning. Among various tensor models, low-Tucker-rank tensors are particularly attractive for capturing multi-mode subspace structures in high-dimensional data. 
Existing recovery methods either operate on the full tensor variable with expensive tensor projections, or adopt factorized formulations that still rely on full-gradient computations, while most stochastic factorized approaches are restricted to tensor decomposition settings.
In this work, we propose a stochastic alternating minimization algorithm that operates directly on the core tensor and factor matrices under a Tucker factorization. The proposed method avoids repeated tensor projections and enables efficient mini-batch updates on low-dimensional tensor factors. Numerical experiments on synthetic tensor sensing demonstrate that the proposed algorithm exhibits favorable convergence behavior in wall-clock time compared with representative stochastic tensor recovery baselines.

\end{abstract}


\IEEEpeerreviewmaketitle

\section{Introduction}
\label{sec:intr}

\IEEEPARstart{T}{ensors} are higher-order extensions of vectors and matrices. They provide a natural and powerful framework for representing and analyzing multi-dimensional data across a wide range of applications in signal processing and machine learning, including hyperspectral imaging~\cite{yang2015compressive,zhao2024robust}, phase imaging~\cite{li2019simultaneous,wang2020tensor}, video processing~\cite{bengua2017efficient,hu2016twist}, wireless communications~\cite{chen2021tensor,de2016overview}, deep networks~\cite{saragadam2024deeptensor,sun2020deep}, and quantum state tomography~\cite{lidiak2022quantum,sofi2025tensor}. However, the extremely high dimensionality of tensor data poses significant challenges in storage, computation, and learning. Tensor decomposition offers an effective tool to exploit the intrinsic low-dimensional structure underlying high-dimensional tensors. Among various models, the CANDECOMP/PARAFAC (CP), Tucker, and tensor train (TT) decompositions are the three most widely used approaches, enabling compact representations and efficient processing of large-scale tensor data~\cite{sokolnikoff1964tensor,kolda2009tensor,sidiropoulos2017tensor}.

In recent years, low-rank tensor recovery from incomplete and linear measurements has attracted significant attention. To address these problems, a variety of recovery algorithms have been proposed from different perspectives. One representative class of methods is based on tensor nuclear norm minimization~\cite{yuan2016tensor,zhang2019corrected,su2019low,li2022nonlinear}. Unlike low-rank matrix recovery, where nuclear norm minimization provides a tractable convex surrogate for rank minimization, computing the nuclear norm of high-order tensors is known to be NP-hard~\cite{hillar2013most,friedland2018nuclear}. Another important class of methods is tensor iterative hard thresholding (TIHT) and its variants~\cite{goulart2015iterative,rauhut2017low,de2017low,grotheer2020stochastic,grotheer2022iterative}. These approaches formulate the recovery problem directly over the full tensor variable and typically require an expensive tensor projection step, whose computational complexity grows rapidly with the tensor dimension. To alleviate these limitations, a third class of methods adopts a factorized formulation and directly optimizes over tensor factors. By representing the target tensor in terms of low-dimensional tensor factors, these approaches significantly reduce the number of optimization variables and avoid explicit tensor projections, leading to improved computational efficiency~\cite{cai2019nonconvex,hao2021sparse,tong2022scaling,qin2024guaranteed,qin2025scalable}.

However, a critical limitation remains: computing full gradients or processing all measurements in each iteration becomes prohibitively expensive for large-scale problems. Stochastic and mini-batch optimization techniques therefore provide a natural way to improve scalability. 
In particular, stochastic optimization has been extensively studied for tensor decompositions from complete data, especially under CP and Tucker factorized models~\cite{vu2015new,ge2015escaping,kolda2020stochastic,anaissi2020online,maehara2016expected,li2020sgd,ahmadi2021randomized}. 
On the other hand, existing stochastic methods for low-rank tensor recovery, such as stochastic TIHT (StoTIHT)~\cite{grotheer2020stochastic}, still operate on the full tensor variable and require expensive tensor projection steps at each iteration, which significantly limits their practical efficiency.

While CP factorization provides a compact representation, Tucker models offer greater flexibility in capturing multi-mode subspace structures and have been widely adopted in tensor sensing and recovery problems. Motivated by these observations, we focus on low-Tucker-rank tensor recovery and investigate whether stochastic optimization can be effectively integrated with Tucker factorization to achieve efficient, projection-free, and scalable recovery from linear measurements.

In this work, we propose a stochastic Tucker alternating minimization (StoTAM) algorithm for Tucker-structured tensor sensing. By explicitly exploiting the Tucker factorization, StoTAM optimizes over the core tensor and factor matrices in an alternating minimization framework. The core tensor is updated via a closed-form least-squares solution, while the factor matrices are updated using mini-batch stochastic gradients followed by QR-based retractions. Compared with the existing StoTIHT method, StoTAM directly optimizes over the tensor factors and avoids repeated tensor projections, which significantly reduces memory and computational costs, making it suitable for large-scale tensor recovery problems. Numerical experiments demonstrate that StoTAM consistently achieves faster convergence in wall-clock time than the StoTIHT baseline.

The remainder of this paper is organized as follows.
In Section~\ref{sec:prel}, we introduce some basic tensor preliminaries used throughout the paper. In Section~\ref{sec:prob}, we formulate the tensor recovery problem and review the StoTIHT baseline. We present the proposed StoTAM algorithm in Section~\ref{sec:algo} and provide numerical simulation results to demonstrate the effectiveness of the proposed method in Section~\ref{sec:simu}. Finally, we conclude the paper and discuss some future directions in Section~\ref{sec:conc}.

\textbf{Notation:} Scalars, vectors, matrices, and tensors are denoted by regular letters (e.g., $x$), bold lowercase letters (e.g., $\vx$), bold uppercase letters (e.g., $\mX$), and bold calligraphic letters (e.g., $\tX$ ), respectively. 
We define $[m]\triangleq\{1,2, \ldots, m\}$. The operator $\Vec(\cdot)$ denotes the vectorization of a matrix or tensor obtained by stacking all its entries into a column vector. The Kronecker product is denoted by $\otimes$. For a vector or matrix, the superscript ``$^\top$'' denotes its transpose. We use $\mX^\dagger$ to denote the Moore-Penrose pseudoinverse of a matrix $\mX$. The identity matrix of size $r$ is denoted by $\I_r$. The Stiefel manifold $\operatorname{St}\left(n, r\right)$ is defined as the set of all $n \times r$ matrices with orthonormal columns, i.e.,
$
\operatorname{St}\left(n, r\right)=\left\{\mU \in \RR^{n \times r}: \mU^{\top} \mU=\I_{r}\right\} .
$

\section{Preliminaries}
\label{sec:prel}
In this section, we briefly review the fundamental concepts and notations from tensor analysis that will be used throughout the paper.
Although we focus on third-order tensors for notational simplicity, all developments can be directly extended to higher-order tensors.

For a third-order tensor $\tX \in \RR^{n_1 \times n_2 \times n_3}$, we use $\calX(i_1, i_2, i_3)$ to denote its $(i_1, i_2, i_3)$-th entry. 
The inner product between two  tensors $\tX, \tY \in \RR^{n_1 \times n_2 \times n_3}$ is
defined as
$$
\langle\tX, \tY\rangle\triangleq\sum_{i_1=1}^{n_1} \sum_{i_2=1}^{n_2} \sum_{i_3=1}^{n_3} \calX(i_1, i_2 ,i_3) \calY(i_1, i_2, i_3).
$$
The induced Frobenius norm is then defined as  $\|\tX\|_F\triangleq\sqrt{\langle\tX, \tX\rangle}$.
These definitions extend the standard inner product and Frobenius norm for vectors and matrices.

Denote $\mX^{(k)}$ as the mode-$k$ matricization/unfolding of a tensor $\tX \in \RR^{n_1 \times n_2 \times n_3}$.  
We say a third-order tensor $\tX \in \RR^{n_1 \times n_2 \times n_3}$ admits a Tucker decomposition with multilinear $\operatorname{rank} \vr=\left(r_1, r_2, r_3\right)$ if it can be written as
$$
\tX=\tS \times_1 \mU_1 \times_2 \mU_2 \times_3 \mU_3,
$$
where $\tS \in \mathbb{R}^{r_1 \times r_2 \times r_3}$ with $r_k = \operatorname{rank} (\mX^{(k)})$ is the core tensor, and the factor matrices $\mU_k \in \mathbb{R}^{n_k \times r_k}$ have orthonormal columns. The product $\times_k$ denotes the mode-$k$ product of a tensor with a matrix. 
We  use $\operatorname{rank}(\tX) \preceq \vr$ to indicate $\operatorname{rank}(\mX^{(k)}) \leq r_k$ for all $k=1,2,3$.

\section{Problem Formulation}
\label{sec:prob}

Consider the recovery of a third-order tensor $\tX^\star \in \RR^{n_1 \times n_2 \times n_3}$ from a set of linear measurements $\vy = \calA(\tX^\star)\in \RR^m$. Here, the linear operator $\calA:\RR^{n_1\times n_2\times n_3}\rightarrow\RR^m$ is defined by
$$
[\calA(\tX)]_i=\langle \tA_i,\tX\rangle,\quad i\in[m],
$$
 where $\tA_i \in \RR^{n_1 \times n_2 \times n_3}$ denotes the $i$-th sensing tensor.
We assume that $\tX^\star$ admits a Tucker decomposition with multilinear rank $\vr = (r_1, r_2, r_3)$, i.e., 
\begin{align}
 \tX^\star = \tS^\star \times_1 \mU_1^\star \times_2 \mU_2^\star \times_3 \mU_3^\star, 
 \label{eq:Xs_decomp}
\end{align}
where $\tS^\star \in \RR^{r_1 \times r_2 \times r_3}$ is the core tensor and $\{\mU_k^\star\}_{k=1}^3 \in \RR^{n_k \times r_k}$ are the factor matrices with orthonormal columns.

In tensor sensing, a commonly used objective is the least-squares loss function
\begin{align*}
    F(\tX) = \frac{1}{2m}\|\vy-\calA(\tX)\|_2^2
= \frac{1}{2m}\sum_{i=1}^m(\langle \tA_i,\tX\rangle-y_i)^2. 
\end{align*}
Given the linear measurements $\vy$ and sensing tensors $\{\tA_i\}_{i=1}^m$, a standard approach to recover $\tX^\star$ is to solve the following problem
\begin{align*}
    \min_{\tX\in \RR^{n_1 \times n_2 \times n_3}} F(\tX) \quad \st \text{rank}(\tX) \preceq \vr.
\end{align*}

By partitioning the measurement vector $\vy$ into $M = \lceil m / b\rceil$ disjoint mini-batches $\{\vy_{[i]}\}_{i=1}^M$\footnote{Note that we use the scalar $y_i$ to denote the $i$-th individual measurement, while $\mathbf{y}_{[i]}$ denotes the measurement vector corresponding to the $i$-th mini-batch.
} of size $b$, the objective function $F(\tX)$ can be rewritten as
\begin{align*}
    F(\tX) = \frac 1 M \sum_{i=1}^M f_i(\tX), 
\end{align*}
where 
\begin{align*}
    f_i(\tX) = \frac{1}{2b} \sum_{j=(i-1)b+1}^{ib} (\langle \tA_j,\tX\rangle-y_j)^2.
\end{align*}

The recovery problem is then naturally cast into a finite-sum form, which is amenable to stochastic and mini-batch optimization methods. A common strategy is to apply stochastic projected gradient updates combined with low-rank tensor projections.
Inspired by IHT methods for compressed sensing and low-rank matrix recovery, as well as TIHT for low-rank tensor recovery, a recent work~\cite{grotheer2020stochastic} proposed the stochastic TIHT (StoTIHT) algorithm, which updates the tensor variable as
$$\tX^{t+1} = \calH_\vr(\tX^t-\mu \nabla f_{i_t}(\tX^t))$$
at each iteration. Here, $\calH_\vr(\cdot)$ denotes a rank-$\vr$ Tucker approximation, typically computed by higher-order singular value decomposition (HOSVD). $\mu$ is the stepsize and $i_t$ is an index randomly seleted from $[M] = \{1,2,\cdots,M \}$ at the $t$-th iteration.

Although StoTIHT is computationally more efficient than full-gradient methods, the repeated tensor projection step can still be expensive for large-scale tensors. Moreover, the theoretical analysis relies on nontrivial assumptions on the approximation quality of $\calH_\vr(\cdot)$.
In addition, the gradient step in StoTIHT is performed directly on the full tensor variable, whose dimension grows exponentially with the tensor order, leading to high memory and computational costs when handling large-scale or high-order tensors.

A natural alternative is to exploit the underlying low-Tucker-rank structure and optimize over the factor matrices and the core tensor instead of the full tensor variable. Such a factorized representation significantly reduces the number of optimization variables, enables more efficient updates, and avoids explicit tensor projection at each iteration.
This motivates us to consider a factorized formulation, which serves as the foundation of the proposed StoTAM algorithm.

Specifically, substituting the Tucker decomposition, i.e., $
\tX = \tS \times_1 \mU_1 \times_2 \mU_2 \times_3 \mU_3$ into the objective function, we get the following factorized optimization problem
\begin{align}
    \min_{\tS;\{\mU_k\}_{k=1}^3\in \text{St}(n_k, r_k)} \!\!G(\tS,\{\mU_k\}_{k=1}^3) \!=\! \frac 1 M \!\sum_{i=1}^M g_i(\tS,\{\mU_k\}_{k=1}^3),
    \label{eq:LRTR_fac}
\end{align}
where $G(\tS,\{\mU_k\}_{k=1}^3) = F(\tS \times_1 \mU_1 \times_2 \mU_2 \times_3 \mU_3)$ and $g_i(\tS,\{\mU_k\}_{k=1}^3) =  f_i(\tS \times_1 \mU_1 \times_2 \mU_2 \times_3 \mU_3)$.

This formulation significantly reduces memory and computational costs, but introduces nonconvexity and coupling among variables.
Our goal is to design a scalable stochastic algorithm that efficiently exploits the factorized structure for solving~\eqref{eq:LRTR_fac}.

\section{The StoTAM Algorithm}
\label{sec:algo}

We now present the proposed stochastic Tucker alternating minimization (StoTAM) algorithm for solving the factorized formulation in~\eqref{eq:LRTR_fac}. The key idea is to combine an alternating minimization scheme over the tensor factors with stochastic (mini-batch) updates enabled by the finite-sum structure of the objective.
Specifically, at the $t$-iteration, StoTAM randomly selects a mini-batch index $i_t\in [M]$ and performs the following two stages of updates:
\begin{enumerate}
    \item Core tensor update: Fix $\{\mU_k\}_{k=1}^3$ and update $\tS$ via a least-squares solution.
    \item Factor matrix updates: Fix $\tS$ and update $\{\mU_k\}_{k=1}^3$  by stochastic gradient steps on the Stiefel manifolds.
\end{enumerate}
This alternating structure avoids explicit tensor projections at each iteration and directly operates on low-dimensional factors, leading to significant computational and memory savings. We next describe the two stages in detail.

\subsection{Core Tensor Update}
In this and the next subsection, for notational simplicity, we occasionally drop the iteration index when no ambiguity arises.
Fixing $\{\mU_k\}_{k=1}^3$, the subproblem with respect to $\tS$ reduces to minimizing the mini-batch objective 
\begin{align*}
    \min_{\tS}~~ g_{i_t}(\tS,\{\mU_k\}_{k=1}^3) = f_{i_t}(\tS \times_1 \mU_1 \times_2 \mU_2 \times_3 \mU_3).
\end{align*}
Define $\mU_{\mathrm{kron}} \triangleq \mU_3 \otimes \mU_2 \otimes \mU_1$ and let $\vs = \Vec{(\tS)}$. The above mini-batch objective can be written as a least-squares function
$$g_{i_t}(\vs) = \frac{1}{2b}\| \mA_{[i_t]}\mU_{\mathrm{kron}} \vs - \vy_{[i_t]}\|_2^2, $$
where $\mA_{[i_t]}\in\RR^{b \times n_1n_2n_3}$ stacks the vectorized sensing tensors in the $i_t$-th mini-batch.
Therefore, the core tensor update admits the following closed-form least-squares solution:
$$\vs^{t+1} = (\mA_{[i_t]}\mU_{\mathrm{kron}}^t)^\dagger \vy_{[i_t]}, $$
which is then reshaped into tensor form to obtain $\tS^{t+1}$.

\subsection{Factor Matrix Updates}

Fixing $\tS$, the factor matrices $\mU_k$ are updated by minimizing $g_{i_t}(\tS,\{\mU_k\}_{k=1}^3)$ with respect to $\mU_k$ on the Stiefel manifold.
The gradient with respect to $\mU_k$ can be derived via standard tensor unfolding and Kronecker product identities. Denote $\vu_k = \Vec{(\mU_k)}$. With some fundamental calculations, we obtain 
\begin{align*}
    \nabla_{\vu_k} g_{i_t}(\tS,\{\mU_k\}_{k=1}^3) = \frac{1}{b} \mOmega_{k_{[i_t]}}(\mA_{[i_t]}\mU_{\mathrm{kron}} \vs - \vy_{[i_t]}),
\end{align*}
where $\mOmega_{k_{[i_t]}} \in \RR^{n_k r_k \times b}$ is defined column-wise as 
\begin{align*}
    \mOmega_{1_{[i_t]}}(:,j) &= \Vec{(\mA_j^{(1)} (\mU_3 \otimes \mU_2) {\mS^{(1)}}^\top )},\\
    \mOmega_{2_{[i_t]}}(:,j) &= \Vec{(\mA_j^{(2)} (\mU_3 \otimes \mU_1) {\mS^{(2)}}^\top )},\\
    \mOmega_{3_{[i_t]}}(:,j) &= \Vec{(\mA_j^{(3)} (\mU_2 \otimes \mU_1) {\mS^{(3)}}^\top )},
\end{align*}
for $j=(i_t-1)b+1, \ldots, i_t b$. $\mA_j^{(k)}$ denotes the mode-$k$ matricization/unfolding of the $j$-th sensing tensor $\tA_j$. Similarly, $\mS^{(k)}$ denotes the mode-$k$ matricization/unfolding of the core tensor $\tS$.

The StoTAM algorithm then performs a stochastic gradient step
\begin{align*}
    \vu_k^{t+1} = \vu_k^t - \mu_{\vu_k} \nabla_{\vu_k} g_{i_t}(\tS^{t+1},\{\mU_\ell^t\}_{\ell=1}^3) ,
\end{align*}
reshapes $\vu_k^{t+1}$ into an $n_k \times r_k$ matrix, and applies a QR-based retraction to enforce orthogonality.
Here, $\mu_{\vu_k}$ is the stepsize.

\subsection{Spectral Initialization.}

To obtain a reliable starting point for the nonconvex optimization in \eqref{eq:LRTR_fac}, we adopt a standard spectral initialization scheme commonly used in low-rank tensor sensing~\cite{qin2025scalable,liang2025landscape}. Specifically, we first form the proxy tensor
$$
\tZ = \frac{1}{m}\calA^\ast(\vy) = \frac{1}{m}\sum_{i=1}^m y_i\tA_i,$$
where $\mathcal A^\ast$ denotes the adjoint operator of $\calA$.
 We then compute a truncated HOSVD of $\tZ$  with target Tucker rank $\vr$ to obtain the initial factor matrices $\{\mU_k^0\}_{k=1}^3$ and the core tensor $\tS^0$. 
 
We summarize the proposed StoTAM algorithm in Algorithm~\ref{alg:stotam}. 

\begin{Remark}
    The proposed StoTAM algorithm (1)  avoids explicit tensor projections at each iteration via a factorized Tucker representation;
(2) updates the core tensor by a closed-form mini-batch least-squares step; and
(3) updates the factor matrices on Stiefel manifolds, leading to efficient stochastic gradient-type updates. As a result, StoTAM significantly reduces the computational burden compared with representative stochastic tensor recovery baselines, especially for high-dimensional tensor recovery problems.
\end{Remark}

\begin{algorithm}[t]
\caption{The StoTAM Algorithm}\label{alg:stotam}
\begin{algorithmic}[1]
\State \textbf{Inputs:} sensing tensors $\{\tA_j\}_{j=1}^m$, measurements $\vy\in\mathbb R^m$, Tucker rank $\vr$, batch size $b$, stepsizes $\{\mu_{\vu_k}\}_{k=1}^3$, total iterations $T$.
\State \textbf{Initialization:} $(\tS^0,\{\mU_k^0\}_{k=1}^3)$ obtained via spectral initialization.
\For{$t=0,1,\ldots,T-1$}
    \State Randomly select a mini-batch index $i_t\in[M]$ and form $(\mA_{[i_t]},\vy_{[i_t]})$.
    \State \textbf{Core update:} set $\mU_{\mathrm{kron}}^t = \mU_3^t\otimes \mU_2^t\otimes \mU_1^t$, $\vs^t= \Vec{(\tS^t)}$.
    \State \hspace{1.1em} $\vs^{t+1} \leftarrow (\mA_{[i_t]}\mU_{\mathrm{kron}}^t)^\dagger \vy_{[i_t]}$, \quad $\tS^{t+1}\leftarrow \mathrm{reshape}(\vs^{t+1})$.
        \State \textbf{Factor updates:} for $k=1,2,3$, let $\vu_k^t= \mathrm{vec}(\mU_k^t)$.
        \State \hspace{1.1em} $\vu_k^{t+1} \leftarrow \vu_k^t - \mu_{\vu_k}\nabla_{\vu_k} g_{i_t}\!\left(\tS^{t+1},\{\mU_\ell^t\}_{\ell=1}^3\right)$.
        \State \hspace{1.1em} Reshape $\vu_k^{t+1}$ into $\widetilde{\mU}_k^{t+1}\in\mathbb R^{n_k\times r_k}$.
        \State \hspace{1.1em} QR retraction: $[\mU_k^{t+1},\sim] \leftarrow \mathrm{qr}(\widetilde{\mU}_k^{t+1},0)$.
        
\EndFor
\State \textbf{Output:} $\widehat{\tX}=\tS^T\times_1 \mU_1^T\times_2 \mU_2^T\times_3 \mU_3^T$.
\end{algorithmic}
\end{algorithm}

\section{Simulation Results}
\label{sec:simu}

In this section, we evaluate the performance of the proposed StoTAM algorithm and compare it with the StoTIHT method~\cite{grotheer2020stochastic}. All experiments are conducted on a MacBook Pro (16-inch, 2023) equipped with an Apple M2 Max chip and 96GB memory, using MATLAB R2023a.

We consider the recovery of a third-order tensor $\tX^\star \in \mathbb{R}^{10\times 10\times 15}$ with Tucker rank 
$\mathbf r=(2,2,2)$. 
The core tensor $\tS^\star$ is generated with i.i.d. standard Gaussian entries, and the factor matrices $\{\mU_k^\star\}_{k=1}^3$ are generated in the same manner and then orthonormalized. The ground-truth tensor is then constructed according to~\eqref{eq:Xs_decomp}.
The sensing tensors $\{\tA_i\}_{i=1}^m$ are generated with i.i.d.\ Gaussian entries following $\calN(0,1/m)$. The measurements are obtained according to the measurement model introduced in Section~\ref{sec:prob}.
We set $m=400$ and partition the measurements into $M=10$ disjoint mini-batches with batch size $b=40$. Spectral initialization is employed for both StoTIHT and StoTAM, and all stepsizes are fixed to 25.

\begin{figure}[htb!]
\centering
\includegraphics[scale=0.475]{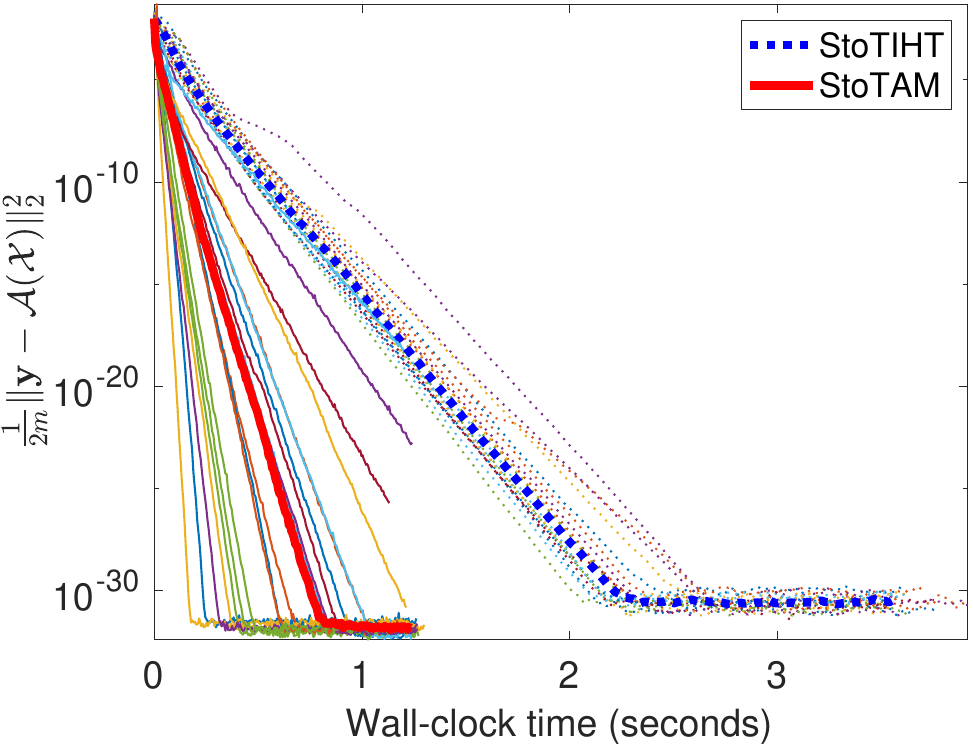} 
\centering
\includegraphics[scale=0.475]{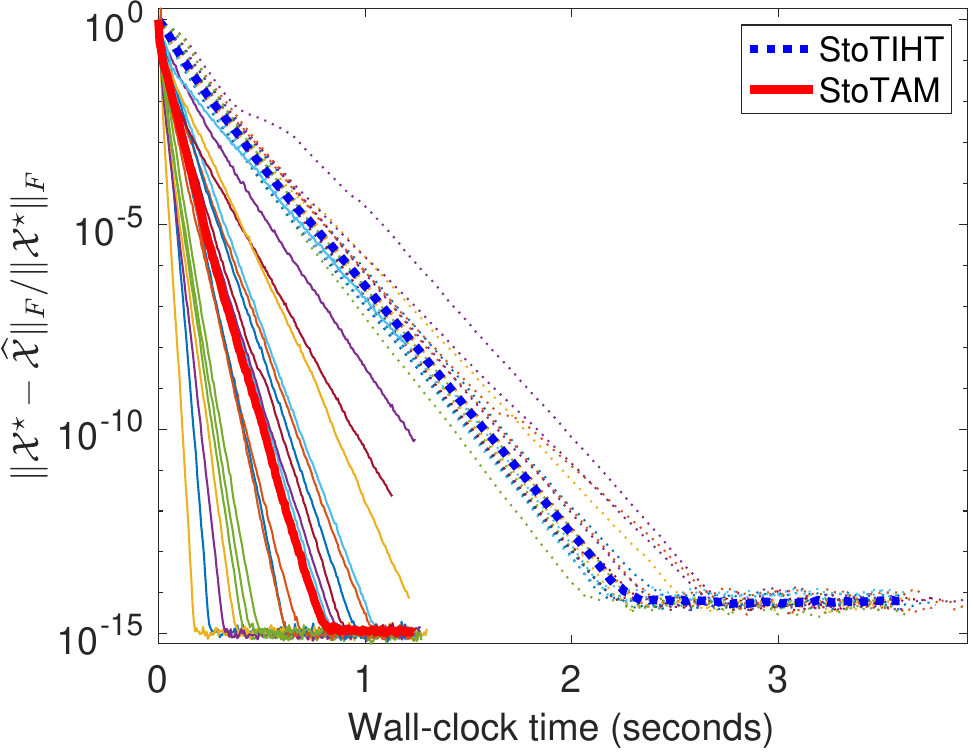} 
\caption{Performance comparison in terms of wall-clock time. Top: loss function value. Bottom: relative reconstruction error. Thin lines represent individual trials (20 trials per algorithm), while thick lines indicate the median performance. In addition, solid lines correspond to StoTAM, and dashed lines correspond to StoTIHT. }\label{fig:loss_errX_time}
\end{figure}

We present the objective function
and the relative recovery error
$
\frac{\|\tX^\star-\widehat{\tX}\|_F}{\|\tX^\star\|_F}$ with respect to wall-clock time over 20 Monte Carlo trials in Fig.~\ref{fig:loss_errX_time}. For each algorithm, we plot all trial trajectories together with their median curves.
As shown in Fig.~\ref{fig:loss_errX_time}, our proposed StoTAM algorithm consistently achieves a much faster decrease in both the loss value and the reconstruction error compared with StoTIHT when measured in wall-clock time. In particular, StoTAM reaches the numerical accuracy floor within approximately one second, whereas StoTIHT requires significantly longer time to attain a comparable accuracy level. Moreover, the median trajectory of StoTAM exhibits a faster initial decrease, indicating more rapid convergence in terms of wall-clock time. These results demonstrate that StoTAM provides substantially improved computational efficiency over StoTIHT, especially in terms of practical runtime performance.

\section{Conclusion}
\label{sec:conc}
This paper proposed a stochastic Tucker alternating minimization (StoTAM) algorithm for low-Tucker-rank tensor recovery from linear measurements. By directly optimizing over the tensor factors, the proposed method avoids repeated tensor projections and significantly reduces both computational and memory costs. The core tensor is updated through a closed-form mini-batch least-squares solution, while the factor matrices are updated through stochastic gradient steps on Stiefel manifolds with orthogonality-preserving retractions. In addition, a spectral initialization scheme is adopted to provide a reliable starting point and enhance the stability of the algorithm. Numerical experiments demonstrate that StoTAM consistently achieves significantly faster convergence in wall-clock time compared with the stochastic tensor iterative hard thresholding (StoTIHT) method. Future work will focus on developing theoretical convergence guarantees, extending the proposed framework to noisy settings and more general structured low-rank tensor models, and exploring applications in large-scale real-world tensor data analysis.

\newpage
\balance
\bibliographystyle{ieeetr}
\bibliography{reference}

\end{document}